\documentclass[letterpaper, 10 pt, journal, twoside]{IEEEtran}
\usepackage[left=16.9mm,right=16.9mm,top=20.1mm,bottom=15.2mm]{geometry}

\usepackage{afterpage}

\usepackage{amsmath,amsfonts}
\usepackage{algorithm}
\usepackage{array}
\usepackage{graphicx}
\usepackage[caption=false,font=small,labelfont=sf,textfont=sf]{subfig}
\usepackage{textcomp}
\usepackage{stfloats}
\usepackage{url}
\usepackage{verbatim}
\usepackage{graphicx}
\usepackage{cite}
\usepackage{xcolor}
\usepackage{soul}
\usepackage{svg}
\usepackage{multirow}
\usepackage{siunitx}
\usepackage{soul}
\usepackage{fancyhdr}
\usepackage{hyperref}
\usepackage{algpseudocode}
\usepackage{booktabs}
\usepackage{multirow}
\usepackage{siunitx}
\usepackage{subcaption}
\usepackage{acro}
\usepackage{svg}

\newcommand{\newac}[2]{\DeclareAcronym{#1}{short=#1,long=#2}}
\newac{FOV}{Field of View}
\newac{IoC}{Intersection over Union}
\newac{PA}{Pixel Accuracy}
\newac{LWIR}{Long-Wave Infrared}
\newac{DEM}{Digital Elevation Map}
\newac{GELU}{Gaussian Error Linear Unit}
\newac{GNC}{Guidance{,} Navigation{,} and Control}
\newac{GPU}{Graphics Processing Unit}
\newac{LRU}{Lightweight Rover Unit}
\newac{PMC}{Pinhole Camera Model}
\newac{RGB-D-T}{RGB{,} depth{,} and thermal}
\DeclareAcronym{LAENTIEC}{short=LAENTIEC,long=Area of Experimentation in New Technologies for Emergencies,foreign=Laboratorio y Area de Experimentación en Nuevas Tecnologías para la Intervención en Emergencias}

\begin{document}

\title{PRISM: Multimodal Terrain Mapping for Rover Navigation in Unstructured Environments}

\author{R.~Castilla-Arquillo$^{1,2}$, C.~J.~Pérez-del-Pulgar$^{2}$, L.~Gerdes$^{2}$, A.~Garcia-Cerezo$^{2}$, and M. Olivares-Mendez$^{1}$
\thanks{$^{1}$R. Castilla-Arquillo and M. Olivares-Mendez are with the Space Robotics (SpaceR) Research Group, SnT, University of Luxembourg, Luxembourg, Luxembourg (e-mail:raul.castilla@uni.lu).}
\thanks{$^{2}$R. Castilla-Arquillo, C. J. Pérez-del-Pulgar, L.~Gerdes, and A. García Cerezo are with the Department of Automation and Systems Engineering, Universidad de Málaga, Andalucía Tech, 29070 Málaga, Spain (carlosperez@uma.es).}}

\maketitle
\begin{abstract}
Robotic navigation in unstructured environments requires robust situational awareness to safely traverse hazards such as steep slopes and rocky terrain. To address this challenge, perception systems increasingly rely on multimodal sensor fusion. Specifically, integrating thermal imagery with standard optical and depth sensors enhances terrain differentiation, directly improving the reliability of mapping algorithms. This paper presents PRISM, a multimodal perception system for terrain mapping in unstructured settings. PRISM leverages a custom sensor suite to capture aligned RGB, depth, and thermal (RGB-D-T) imagery. At its core is OmniUnet, a novel vision transformer-based network specifically designed for multimodal semantic terrain segmentation. We validated the proposed system using two newly annotated datasets (BASEPROD and LAENTIEC) and demonstrate its real-world applicability through physical field experiments. Deployed on a resource-constrained embedded computer, PRISM efficiently generates traversability maps that directly enable autonomous navigation via a rover’s Guidance, Navigation, and Control (GNC) subsystem.
\end{abstract}

\section{INTRODUCTION} 
\label{sec:introduction}
Advancements in rover autonomy, driven by the Guidance, Navigation, and Control (GNC) system, have enabled reliable operation in unstructured and challenging terrains where direct human control is limited by communication constraints or environmental complexity \cite{quadrelli2015guidance}. Such environments often feature hazards like steep slopes, dense vegetation, and uneven ground, which can lead to slippage or entrapment \cite{Wijayathunga2023, gonzalez2018slippage}. To address these risks, the navigation module of the GNC system relies on terrain mapping generated from onboard sensor data.

The terrain mapping process consists of two main stages: terrain segmentation and traversability analysis \cite{borges2022survey}. In the first stage, segmentation identifies distinct terrain regions and obstacles based on surface features. This is commonly handled by deep learning models for image segmentation, such as U-Net \cite{ronneberger2015u}, Mask R-CNN \cite{he2017mask}, and YOLO \cite{diwan2023object}. More recent approaches incorporate Vision Transformers (ViT) \cite{kolesnikov2021vit} and hybrid architectures, such as Swin-Unet \cite{cao2022swin}, which have demonstrated strong performance in complex environments \cite{liu2023survey, wang2022unetformer}. Most of these segmentation models, however, are designed for a single input modality, which limits their direct applicability when perception must draw on multiple complementary sensors simultaneously. 

During the second stage of terrain mapping, traversability analysis utilizes segmented data to produce a map assessing navigational difficulty. Traversability is generally derived from geometric features, such as slope and roughness extracted from elevation data, or from machine learning models that predict rover behavior based on kinematics, energy use, and slip risk \cite{zhou2019mapping, ono2020maars, gaines2020self}. Because the Navigation module relies on these maps for path planning, the accuracy of these foundational stages dictates the overall safety of autonomous operations.


In the context of multimodal terrain mapping, the specific sensors a perception system fuses depend on the target operating environment. Within structured environments, current research combines RGB with stereoscopic images to segment small obstacles \cite{gupta2018mergenet, hua2019small, sun2020real} and on-road features \cite{zhao2023cross}, while pairing thermal imagery with RGB data has gained traction for on-road elements \cite{vertens2020heatnet, wang2023dacfn, zhang2023cmx, liu2023urtsegnet}. Such fusion overcomes the limitations of any single sensor in complex scenes \cite{zhang2023perception}, a principle reflected in rover sensor suites: the ScienceCam on the Light Weight Rover Unit (LRU) combines grayscale, RGB, hyperspectral, and thermal imaging \cite{Wedler2017}, and Mars rovers such as Curiosity \cite{grotzinger2012mars} and Perseverance \cite{Bell2021} carry similarly diverse suites for navigation and science. Perception in unstructured terrain, by contrast, still relies predominantly on fusing RGB and LiDAR \cite{leung2022hybrid, zhong2022off, kim2024ufo}, and the integration of thermal imaging remains comparatively underexplored \cite{gonzalez2017thermal, iwashita2020virtual}.

Addressing this gap is important, as thermal imaging offers unique insights in unstructured terrain. Surface thermal properties, such as thermal inertia, relate directly to traversability: under solar heating, granular sandy soils reach higher surface temperatures than more compacted soils, providing a cue for distinguishing terrain types that is difficult to recover from RGB data alone \cite{gonzalez2017thermal}. The cue is even stronger on Mars, where low atmospheric pressure heightens thermal contrasts and supports the estimation of rover slippage \cite{cunningham2019improving, castilla2023thermal}. Exploiting it, however, requires two integrated elements: first, an \acs{RGB-D-T} sensing system that captures visible light (RGB), depth (D), and thermal (T) modalities; and second, a model capable of fusing these heterogeneous channels within a single architecture \cite{omniunet2025castilla}.

Several multimodal datasets have been proposed to support navigation in unstructured terrain. RUGD \cite{wigness2019rugd} pairs a monocular camera with LiDAR on a mobile robot to record diverse terrain classes, and RELLIS-3D \cite{jiang2021rellis} provides synchronized stereo vision and LiDAR data with semantic terrain annotations across varied outdoor settings. The CAVS dataset \cite{sharma2022cat} contributes paired image and LiDAR data capturing fluctuations in lighting, weather, and vegetation density, while BASEPROD \cite{gerdes2024baseprod} offers RGB, depth, and thermal data collected by a rover in the Martian-like Bardenas semi-desert of Spain, matching the modality requirements of this work. Few systems, however, integrate thermal sensing into a multimodal segmentation pipeline and validate the resulting traversability map on a rover under real field conditions. Closing this gap, from offline segmentation to a deployed, field-tested mapping system, is the focus of this work.

Building on this, we present PRISM (Perception for RGB, Infrared, and Spatial Mapping), a multimodal terrain mapping system designed to integrate with the Navigation module of the GNC architecture for autonomous rover operations in unstructured environments. The system leverages \ac{RGB-D-T} data to produce traversability maps that account for rover constraints. To process this data, we implement our previously introduced OmniUnet architecture \cite{omniunet2025castilla}, a deep learning model that combines the strengths of the Omnivore and U-Net frameworks. To support this system, we developed a multimodal sensor housing to mount multiple cameras on a robotic platform, the Rover Autonomy Testbed (RAT). We also contribute two publicly available labeled multimodal datasets: a labeled subset of the BASEPROD dataset \cite{gerdes2024baseprod}, collected in the Bardenas semi-desert, and a new dataset gathered during rover traverses at the University of Málaga's Area of Experimentation in New Technologies for Emergencies (LAENTIEC). Both span a variety of terrain types, such as dirt, grass, and gravel. We train and validate the terrain segmentation component, powered by OmniUnet, on both the BASEPROD and LAENTIEC datasets, and we validate the PRISM system through field experiments with the RAT platform in the unstructured terrain of LAENTIEC. The implementation was carried out in ROS2 and deployed on an onboard embedded computer. The source code and datasets are publicly available on GitHub\footnote{\url{https://github.com/spaceuma/MultimodalNavigation}}.

\section{The Guidance, Navigation, and Control system}

The Guidance, Navigation, and Control (GNC) system employed in this work enables autonomous rover operation in complex, unstructured environments by fusing data from multiple sources. Inspired by architectures used in planetary exploration missions, it combines drone-based photogrammetry, which simulates the multispectral satellite data used in those missions, with the rover's onboard sensor data. These include exteroceptive sensors, such as cameras and LiDAR, for environmental perception, and proprioceptive sensors, such as wheel encoders and inertial units, for tracking the rover’s motion. 

\begin{figure}
  \centering
  \includegraphics[width = 0.49 \textwidth]{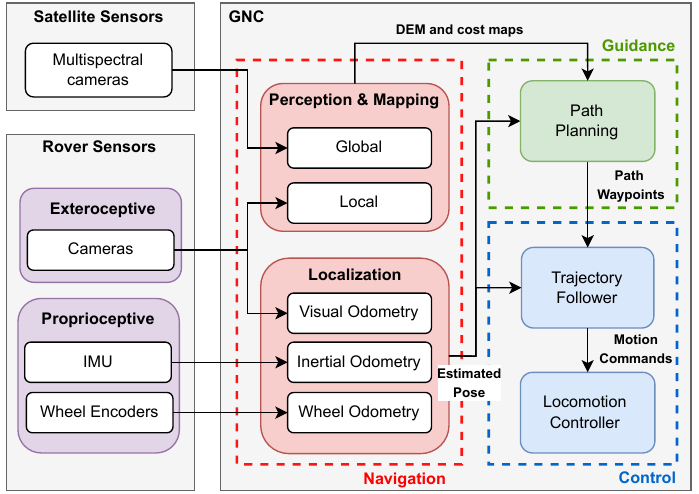}
  \caption {\small Architecture of the Guidance, Navigation, and Control (GNC) system. The Path Planning module receives reference inputs from the operator, while the Locomotion Controller generates actuator command signals.} 
  \label{fig:gnc}
\end{figure}

When an operator assigns a scientific target or point of interest as a goal, the GNC system processes this objective to plan a traversable path based on the current terrain. It then commands the rover’s actuators to follow the generated trajectory, ensuring accurate and safe motion toward the target. As illustrated in Figure~\ref{fig:gnc}, the architecture comprises five main subsystems that collaborate in real time to generate terrain maps, estimate position, plan safe paths, and control movement. This section provides a brief overview of each component, details the specific algorithms used, and provides access to our open-source implementations.

\textbf{Perception \& Mapping (Navigation):} Responsible for building global and local maps to be used for path planning \cite{azkarate2022design}. The global map, obtained from drone photogrammetry, provides a broad, low-resolution view of the terrain and supports route planning over large areas. The local map, continuously updated during traversal using exteroceptive sensors, captures nearby obstacles and fine-scale terrain features not visible in the global data. These maps are generally combined into Digital Elevation Maps (DEMs) and traversability maps \cite{borges2022survey}. A DEM represents the environment as a two-dimensional grid, with each cell storing the elevation at that location. Traversability maps, on the other hand, assign a traversal cost to each cell based on terrain characteristics such as slope or roughness. Both maps are sent to the Path Planning subsystem to support the computation of safe and efficient paths. In this work, the mapping pipeline is driven by our custom global and local DEM layers, which are based on the algorithms presented in \cite{paz2020improving}. The implementations for both layers are available as open-source repositories on GitHub\footnote{Global DEM: \url{https://github.com/spaceuma/nav2_global_dem_layer}; Local DEM: \url{https://github.com/spaceuma/nav2_local_dem_layer}}.
    
\textbf{Localization (Navigation):} Estimates the rover’s pose, including its position, orientation, and velocity, by combining data from exteroceptive and proprioceptive sensors \cite{zarei2022advancements}. Wheel odometry tracks movement using encoder data, while visual odometry relies on camera images to correct for slippage and improve accuracy. Inertial odometry, derived from the Inertial Measurement Unit (IMU), which records acceleration and angular velocity, improves pose estimation by providing reliable motion data during rapid movements or in visually degraded environments. Localization can be integrated with mapping through Simultaneous Localization and Mapping (SLAM), a process in which the system incrementally constructs a map of the environment while simultaneously improving its estimate of the rover’s pose within that map \cite{stachniss2016simultaneous}. The final pose estimate is shared with the Path Planning and Trajectory Follower subsystems. State estimation is achieved using a GNSS/IMU localization node, which is available as an open-source repository\footnote{\url{https://github.com/spaceuma/nav2_gnss_imu_localization_node}}.
    
\textbf{Path Planning (Guidance):} Determines a route to a target location, which may be manually assigned by an operator or selected autonomously based on scientific interest or mission objectives. It takes into account the rover’s current pose and the terrain information provided by the perception and mapping subsystem. Using this data, it generates a sequence of waypoints that avoid obstacles and account for terrain traversability. The planned path aims to minimize risk while optimizing travel distance or energy consumption, depending on the mission constraints \cite{sanchez2021path}. Once computed, the waypoint sequence is passed to the Trajectory Follower for execution. For trajectory generation, we deployed a Fast Marching Method (FMM) planner \cite{sanchez2019dynamic} adapted for unstructured terrain, with the source code publicly accessible on GitHub\footnote{\url{https://github.com/spaceuma/nav2_fmm_multilayered_planner}}.

\textbf{Trajectory Follower (Control):} Converts the planned path into motion commands in real time. It dynamically adjusts the rover’s speed, heading, and orientation to follow the trajectory accurately while responding to terrain conditions and deviations. One of the most widely used algorithms for this task is the Pure Pursuit algorithm \cite{wallace1985first}, which calculates steering commands based on geometric criteria. Key parameters include the look-ahead distance, which defines how far ahead along the path the algorithm targets, and the width of the safety corridor, which sets the maximum allowable deviation from the trajectory before the path is considered unrecoverable. The resulting motion commands are sent to the Locomotion Controller for execution. Specifically, our framework implements Conservative Pure Pursuit Waypoint Navigation with a safety corridor \cite{Gerdes2020}, which is available online\footnote{\url{https://github.com/spaceuma/nav2_conservative_pursuit}}.
    
\textbf{Locomotion Control (Control):} Executes movement commands through the rover’s actuation system, whether using wheels, tracks, or legs. It compensates for terrain disturbances, enforces actuator limits, and applies control strategies such as Proportional-Integral-Derivative (PID) \cite{kim2016kinematic} control or Model Predictive Control (MPC) \cite{he2022model} to ensure stable and accurate locomotion. In wheeled systems, it may also coordinate differential or skid steering and manage suspension dynamics to maintain stability and traction across challenging terrain.

Together, these five subsystems form an integrated GNC architecture that enables robust and adaptive navigation in unstructured environments. Its modular design allows for scalability and adaptability across different robotic platforms and mission scenarios. This foundational structure serves as the basis for integrating advanced perception systems, such as the multimodal terrain-mapping module in this work.

\section{Multimodal Terrain Mapping System}
\label{sec:perception-sistem}

\begin{figure}[t]
    \centering
    \includegraphics[width=0.49\textwidth]{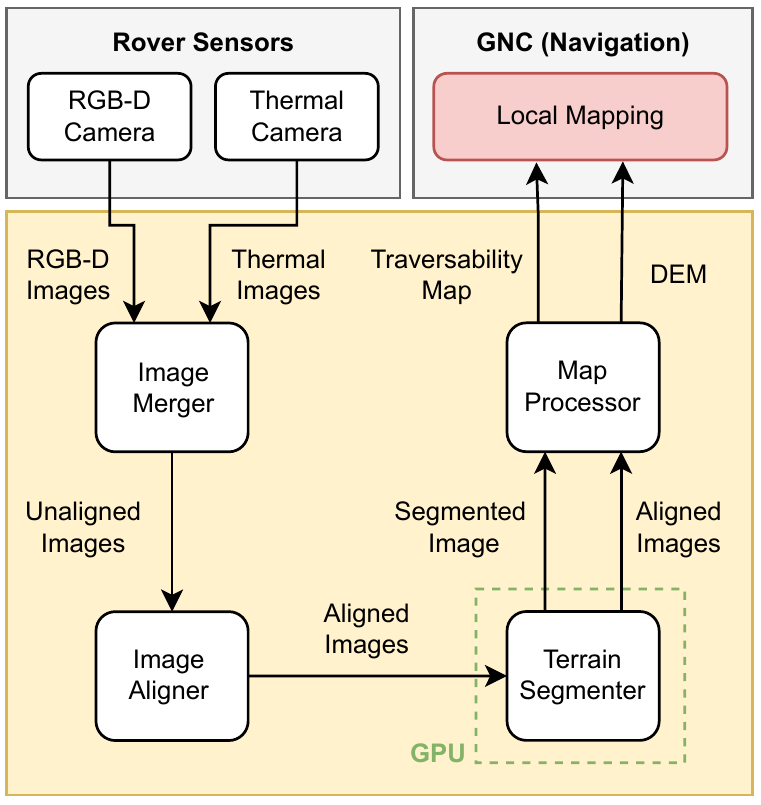}
    \caption{ \small Architecture of the Perception for RGB, Infrared, and Spatial Mapping (PRISM) system.}
    \label{fig:multimodal-system}
\end{figure}

The proposed multimodal terrain mapping system, PRISM, is integrated into the Perception \& Mapping subsystem, which generates local maps using data from the rover’s exteroceptive sensors. These maps provide elevation and traversability information and are forwarded to the path-planning subsystem to ensure safe navigation. As illustrated in Figure~\ref{fig:multimodal-system}, PRISM consists of four interconnected components that operate concurrently: the image merger, image aligner, terrain segmenter, and map processor.


\begin{figure*}
    \centering
    \subfloat[]{\includegraphics[height=0.37\textwidth]{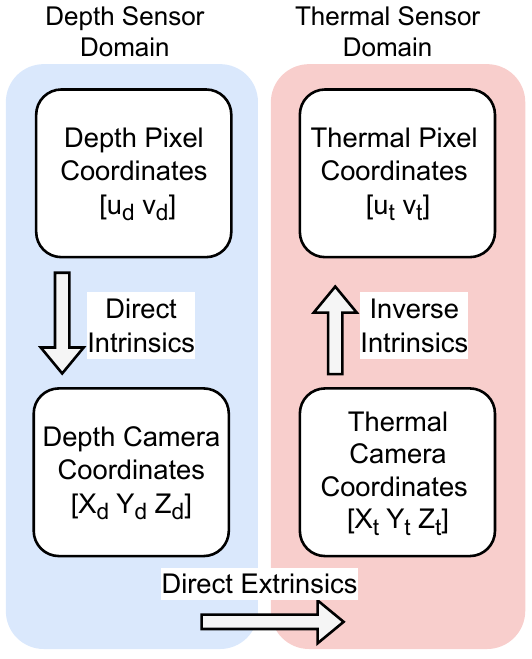}}
    \hspace{0.05cm}
    \subfloat[]{\includegraphics[height=0.37\textwidth]{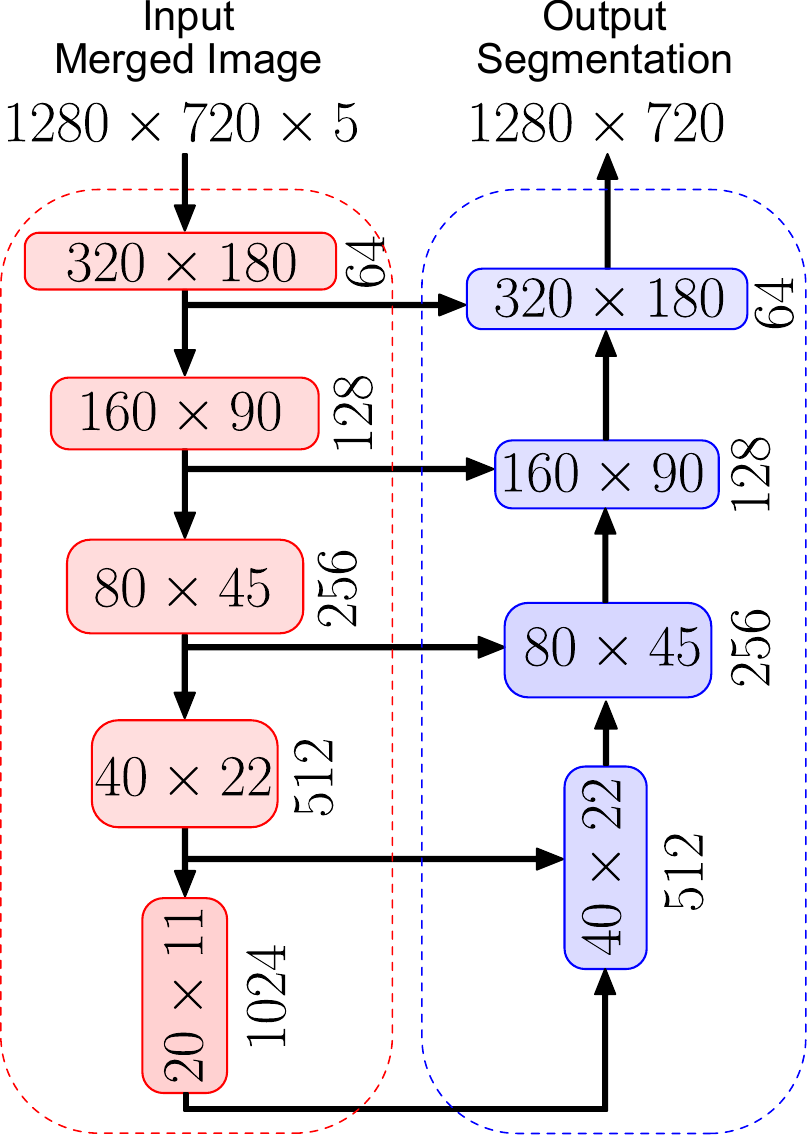}}
    \hspace{0.05cm}
    \subfloat[]{\includegraphics[height=0.37\textwidth]{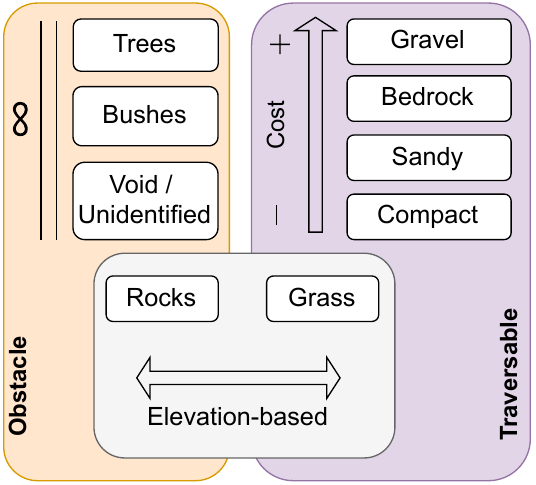}}
    \caption{\small Core pipeline stages of the PRISM system: (a) thermal-to-depth alignment (Image Aligner); (b) OmniUnet-based terrain segmentation (Terrain Segmenter); and (c) terrain- and elevation-based traversability mapping (Map Processor).}
    \label{fig:modules-diagrams}
\end{figure*}

The process begins with the image merger, which integrates the RGB-D and thermal camera images based on their timestamps. These temporally synchronized but spatially unaligned images are then processed by the image aligner, which leverages depth data from the stereo camera to achieve precise spatial alignment. Once aligned, the images are passed to the terrain segmenter, which operates on an embedded \ac{GPU} to generate segmentation masks. These masks, along with the aligned images, are then used by the map processor to construct a \ac{DEM} and a traversability map that incorporate depth and segmentation data, respectively. Finally, both maps are integrated into a multilayered \ac{DEM} forming the rover’s local map. This map is then sent to the rover's \acs{GNC} system to enhance its ability to navigate unstructured terrain.

\subsection{Image Merger}

The process begins by retrieving the latest data from the sensor stacks to initialize a unified multi-channel matrix, $M_{out}$, which is flattened into a one-dimensional array for efficient serialization.  The data is mapped into five channels: channels 0 through 2 contain the RGB data, channel 3 holds the depth information, and channel 4 stores the thermal data. Any out-of-bounds thermal pixels are set to zero. Finally, the rover's current pose and GPS coordinates are appended to the payload, providing spatial context before it is published to the navigation stack.

\subsection{Image Aligner}

Although the stereo camera inherently outputs registered RGB and depth images, the thermal camera requires a separate alignment step to match this coordinate frame. This registration relies on the \ac{PMC}, which mathematically models the projection of a 3D scene onto a 2D image plane. The model uses extrinsic parameters to define the camera's spatial position and orientation, along with intrinsic parameters that specify the focal length and optical center. As shown in Fig.~\ref{fig:modules-diagrams}a, the proposed method aligns the thermal image by adapting standard stereo depth registration techniques. The algorithm initializes by storing the incoming depth and thermal frames as matrices and creating an empty aligned thermal matrix. 

To populate this matrix, the system processes every pixel in the depth sensor domain with coordinates $[u_{d}, v_{d}]$ through a specific sequence of geometric transformations. First, it applies the depth camera's intrinsic parameters to back-project the 2D pixel into 3D depth camera coordinates $[X_{d}, Y_{d}, Z_{d}]$. Next,  extrinsic parameters are applied to transform this 3D point into the thermal camera's coordinate space, yielding $[X_{t}, Y_{t}, Z_{t}]$. Finally, the thermal camera's intrinsic parameters are used to project this 3D point onto the 2D thermal sensor domain, calculating the corresponding thermal pixel coordinates $[u_{t}, v_{t}]$. If these projected coordinates fall within the boundaries of the thermal image frame $F_{T}$, the corresponding temperature value is assigned to that pixel location in $A_{T}$. Out-of-bounds pixels are ignored.

\subsection{OmniUnet terrain segmenter}

The terrain segmenter receives the aligned multimodal images and generates a semantic segmentation mask of the rover's surroundings. In addition to producing this segmented image, the module concurrently forwards the aligned multimodal images to the map processor for further integration. This component is powered by the OmniUnet architecture \cite{omniunet2025castilla}, as depicted in Fig.~\ref{fig:modules-diagrams}b, which is a transformer-based neural network designed to process combined RGB, depth, and thermal imagery. The model uses an Omnivore backbone \cite{girdhar2022omnivore} with shifted-window attention mechanisms to extract features across different sensory modalities, alongside a U-Net decoding strategy to produce the final terrain classifications. As illustrated in the diagram, the red blocks represent the Swin Transformer blocks responsible for feature extraction, while the blue blocks denote the convolutional layers with \ac{GELU} activations used during the decoding stage. The numbers inside these blocks indicate the spatial dimensions of the feature maps, while the numbers outside denote the channel dimensions at each respective stage. To ensure the system remains suitable for autonomous field operations, the segmenter is deployed on an onboard embedded GPU. This hardware configuration enables the network to perform efficient inference when classifying complex, unstructured terrain, such as sand, bedrock, and compact soil.


\subsection{Map processor}

\algnewcommand{\LeftComment}[1]{%
\vspace{1.0em}
\Statex \textbf{#1}:
\vspace{0.3em}}

\begin{algorithm}[t]
    \begin{algorithmic}[1]
        \LeftComment{Initialization}
        \State $\mathit{res} \leftarrow$ obtainMapResolution()
        \State $\mathit{pos} \leftarrow$ getCameraPose()
        \State $F_{D} \leftarrow$ obtainDepthFrame()
        \State $F_{S} \leftarrow$ obtainAlignedSegmentationFrame()
        \State $P_{w} \leftarrow$ initializeWorldPointsArray($F_{D}$)
        \State $P_{s} \leftarrow$ initializeSegmentationPointsArray($F_{S}$)
        \State $n \leftarrow$ 0
        
        \LeftComment{Creating arrays with world points and seg.\ values}
        \For{$[u_{d}, v_{d}]$ \textbf{in} $F_{D}$}
            \State $[X_{w}, Y_{w}, Z_{w}] \leftarrow$ getWorldCoord($F_{D}$, $[u_{d}, v_{d}], \mathit{pos}$)
            \State $P_{w}(n) \leftarrow$ $[X_{w}, Y_{w}, Z_{w}]$
            \State $F_{s}(n) \leftarrow$ $P_{s}\left([u_{d}, v_{d}]\right)$
            \State $n \leftarrow$ $n+1$
        \EndFor
        
     \LeftComment{Setting map bounds}
     \State $x_\mathit{offset} \leftarrow$ $\min(P_{w}[X_{w}])$
     \State $M_\mathit{width} \leftarrow$ $\left(\max(P_{w}[X_{w}]) + x_\mathit{offset}\right) \cdot \mathit{res}$
     \State $M_\mathit{height} \leftarrow$ $\max(P_{w}[Y_{w}])\cdot \mathit{res}$
     \State $M \leftarrow$ initializeMatrix($[M_\mathit{width},M_\mathit{height}]$)
     \State $S \leftarrow$ initializeMatrix($[M_\mathit{width},M_\mathit{height}]$)
     
    \LeftComment{Generating elevation and segmentation maps}
    \For{$n$ \textbf{in} $P_{w}$}
        \State $u_{m} \leftarrow$ $(P_{w}(n)[X_{w}] + x_\mathit{offset}) \cdot \mathit{res}$
        \State $v_{m} \leftarrow$ $P_{w}(n)[Y_{w}] \cdot \mathit{res}$
        \State $M([u_{m},v_{m}]) \leftarrow$ $P_{w}(n)[Z_{w}]$
        \State $S([u_{m},v_{m}]) \leftarrow$ $P_{s}(n)$
    \EndFor
    \end{algorithmic}
    \caption{ \small Computation of elevation ($M$) and segmentation ($S$) maps.}
    \label{alg:dem_computation}
\end{algorithm}

The map processor performs two primary functions. First, it generates a multilayered \ac{DEM} that combines RGB, thermal, elevation, and segmentation data. Second, it creates a traversability map using the terrain classifications from the terrain segmenter and the elevation data from the \ac{DEM}. The multilayered \ac{DEM} is generated following the procedure outlined in Algorithm~\ref{alg:dem_computation}. The system first collects depth and segmentation frames, along with the depth camera's orientation. It then processes the depth frame iteratively, converting each depth value into world coordinates using the camera's intrinsic parameters. This step yields two main arrays: one containing the spatial coordinates and the other containing the terrain classes. The system then evaluates the coordinates to determine the appropriate map dimensions for a specific resolution. Finally, it initializes the elevation and segmentation matrices and populates them with the aligned data.

To construct the traversability map, the system evaluates both the predicted terrain class and the corresponding ground elevation. Figure~\ref{fig:modules-diagrams}c depicts this process and provides examples of different terrain categories. The map classifies the environment into two main groups based on navigation difficulty.

\textbf{Obstacle:} These areas receive an infinite cost regardless of the depth camera measurements. Certain terrains are classified as obstacles if their geometry is too steep or rough for the rover to cross safely. As a safety measure, any object not recognized by the terrain segmenter is automatically treated as an obstacle.

\textbf{Traversable:} These areas are assigned varying costs based on how easily the rover can navigate them. Specific terrain types receive higher costs depending on their height and surface properties.


The costs generated from this process must be configured to meet the requirements of the \ac{GNC} navigation stack. Once established, the traversability map can be further improved by incorporating data from other rover sensors. This combined approach is particularly useful for ambiguous areas such as grass and rocks, where relying solely on elevation or visual data can lead to classification errors. For example, if the system relies exclusively on elevation data, it might interpret a patch of tall grass as a solid obstacle. However, by including the neural network prediction, the system correctly identifies the grass and allows the rover to pass through. Conversely, the segmenter might identify a rock and flag it as an obstacle. By checking the elevation data, the system can determine if the rock is actually low enough for the rover to drive over safely.

\begin{figure*}
    \centering
    \subfloat[]{\includegraphics[height=0.30\textwidth]{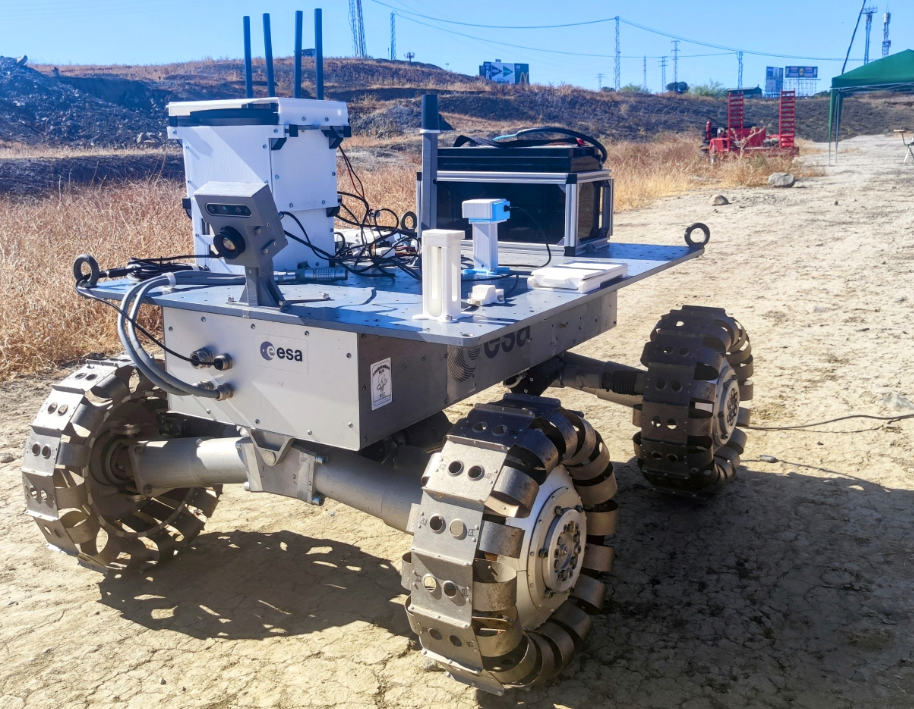}}
    \hspace{0.1cm}
    \subfloat[]{\includegraphics[height=0.30\textwidth]{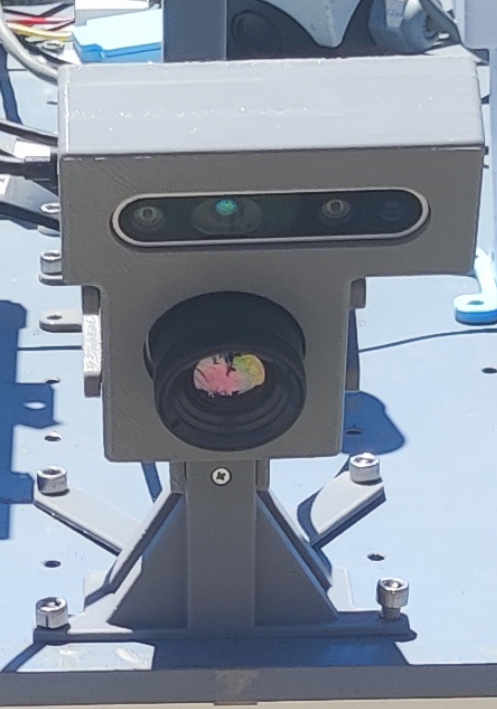}}
    \hspace{0.1cm}
    \subfloat[]{\includegraphics[height=0.30\textwidth]{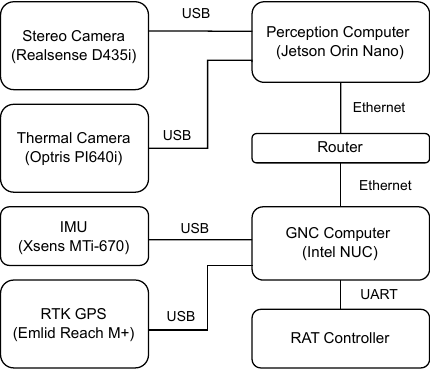}}
    \caption{Hardware configuration: (a) RAT platform during field operations, (b) 3D-printed multimodal sensor housing, and (c) block diagram of the system's hardware architecture.}
    \label{ffig:hardware}
\end{figure*}

\section{Implementation and experiments}

To validate the proposed PRISM architecture, this section details the complete system implementation and the experimental procedures. First, we describe the hardware configuration of the robotic platform and the custom multimodal sensor suite. Next, we introduce the datasets used for training and evaluation, followed by the specific training parameters used for the OmniUnet model. Finally, we present results from field experiments conducted in an unstructured environment to demonstrate the system's real-world applicability and performance.

\subsection{Hardware}

The RAT rover (Figure~\ref{ffig:hardware}a) was used to perform the experiments. Its locomotion system features a Double-Bogie configuration with a passive longitudinal differential and a $4 \times 4 \times 4$ layout, incorporating four wheels, each with independent driving and steering capabilities~\cite{medina2015design}. Additionally, it includes an Xsens MTi-670 IMU and an Emlid Reach M+ GNSS receiver.

Multimodal images were generated by combining data from a Realsense D435i stereo camera and an Optris PI-640i thermal camera. The Realsense RGB sensor produces images with dimensions up to $1920\times1080$ pixels and a \ac{FOV} of \qtyproduct{69x42}{\degree}. It provides stereoscopic depth images at a resolution of $1280\times720$, with a \ac{FOV} of \qtyproduct{87x58}{\degree} and depth accuracy of less than \SI{2}{\percent} at a distance of \SI{2}{m}. The Optris PI-640i thermal camera is based on uncooled microbolometer technology with a resolution of $640\times480$ pixels. It operates in the \ac{LWIR} spectral range from \SIrange{8}{14}{\micro\meter} and can measure temperatures ranging from \SIrange{-20}{900}{\celsius} with a thermal sensitivity of \SI{0.04}{\celsius}. Additionally, it features a germanium optic with a \qtyproduct{60x45}{\degree} \ac{FOV}. 

A custom 3D-printed housing was designed to enclose both cameras (Fig.~\ref{ffig:hardware}b), ensuring that the depth plane of the Realsense camera was vertically aligned with the optical plane of the thermal camera. This alignment enabled accurate projection of thermal data into the 3D world. The multimodal sensor was mounted on the platform at a height of \SI{65}{\cm} with a \ang{20} inclination to capture surface images as comprehensively as possible.

The RAT was equipped with two onboard computers linked via an Ethernet router; a diagram is depicted in Fig.~\ref{ffig:hardware}c. One computer was dedicated to the rover's GNC and the other handled perception. The GNC computer was an Intel NUC with an Intel Core i5 processor running at \SI{1.8}{\giga\hertz} and \SI{16}{\giga\byte} of RAM. This unit received data from the IMU and RTK GPS via USB connections and sent commands to the RAT controller over UART. The GNC architecture was programmed in ROS~2 and used the NAV2 \cite{macenski2020marathon2} stack to execute a goal command, following a trajectory computed with the fast marching algorithm \cite{Sanchez-Ibanez2019}. The perception computer was a Jetson Orin Nano featuring an ARM processor operating at \SI{1.5}{\giga\hertz}, \SI{8}{\giga\byte} of RAM, and an AI performance of 40 TOPS. It connected directly to the stereo and thermal cameras via USB. This computer was chosen over a high-performance GPU due to its low power consumption of \SI{15}{\watt} and its limited resource architecture.

\subsection{Terrain Segmenter Training}

Because there are currently no publicly available multimodal datasets of the Martian surface, multimodal images of unstructured terrestrial environments were utilized to train and evaluate the proposed system. We contribute two publicly available labeled multimodal datasets for this purpose. To ensure complete consistency across all our data and field experiments, we used the same multimodal sensors and calibration parameters as those originally used to capture the BASEPROD dataset. The generated segmentation masks are single-channel images matching the resolution of the captured data, with unique integers representing different terrain classes. RGB images are provided in PNG format, whereas depth and radiometric thermal data are stored as CSV files. 

\textbf{The RUGD Dataset:} The network was initially trained and evaluated using the RUGD dataset to establish a performance baseline. This dataset contains approximately 7,500 labeled RGB images of unstructured outdoor scenes annotated across 25 semantic classes \cite{wigness2019rugd}.

\textbf{The Bardenas Dataset:} The first multimodal dataset builds upon the BASEPROD dataset~\cite{gerdes2024baseprod}, which was collected during traverses with the MaRTA rover \cite{azkarate2022design} in the Bardenas Reales semi-desert located in Cabanillas, Navarra, Spain. This environment features diverse terrains and geological obstacles, such as rocks and exposed bedrock, that closely mimic Martian surface characteristics. While the original BASEPROD dataset contains ca.\ 36,000 raw multimodal images, it does not provide semantic annotations. To address this, we contribute a newly labeled subset of this data. We manually labeled 950 representative images for training and 190 for validation. The annotations encompass eight classes: void, compact, grass, bedrock, sandy, gravel, rock, and bush. We have published these labeled segmentation masks on Zenodo\footnote{\url{https://doi.org/10.5281/zenodo.15496884}} \cite{Castilla2025SegmentationData} to support future research.

\textbf{The LAENTIEC Dataset:} The second multimodal dataset was generated using the RAT rover at LAENTIEC, an unstructured \SI{90000}{\square\meter} testing facility at the University of Malaga, Spain. LAENTIEC provides varying elevations and diverse terrain types including dirt, gravel, and sand. Captured in spring, this dataset features denser green vegetation than the Bardenas dataset, though the site typically dries out in summer. The LAENTIEC dataset~\cite{zenodo2023rgbdt} consists of 310 multimodal images, with 62 reserved for validation. It classifies six semantic classes: void, compact, grass, gravel, sky, and tree/bush. This newly collected dataset has also been made publicly available on Zenodo\footnote{\url{https://doi.org/10.5281/zenodo.8032971}}.

The training was conducted on an Nvidia DGX Station using a GPU Tesla V100 DGXS with \SI{32}{\giga\byte} of V-RAM. The network was trained for $50$ epochs for all modalities with a batch size of $16$ images and a learning rate of \num[output-exponent-marker=\text{e}]{2e-5}. The optimization function was a combined loss comprising Dice loss and cross-entropy loss. The datasets were split into 80/20 training/validation sets. 

\begin{figure*}[t]
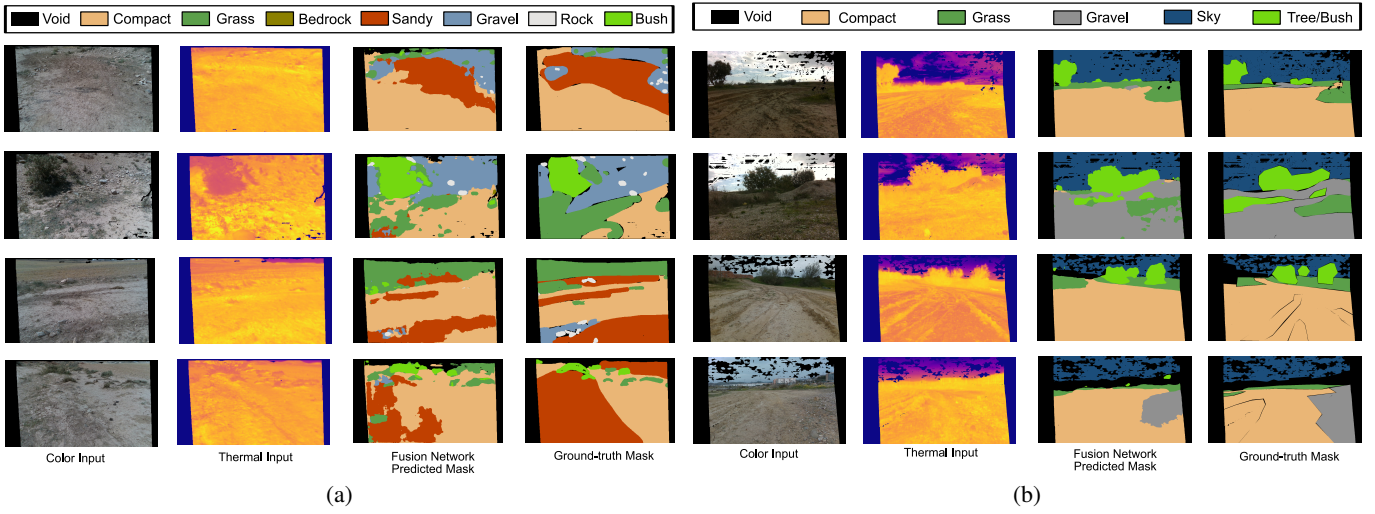

\centering
\captionsetup{width = 1\linewidth}
    \centering
    \subfloat[]{\includesvg[height=0.34\textwidth]{images/omniunet_baseprod_mod.svg}}
    \hspace{0.1cm}
    \subfloat[]{\includesvg[height=0.34\textwidth]{images/omniunet_laentiec_mod.svg}}
\caption{Qualitative comparison of OmniUnet multimodal segmentation. Each subfigure presents the RGB and thermal inputs alongside the predicted and ground-truth segmentation masks for (a) the Mars-analog Bardenas dataset and (b) the LAENTIEC dataset.}
\label{fig:omniunet-evaluation}
\end{figure*}

Due to the limited number of images in the LAENTIEC dataset, OmniUnet was initially pre-trained on \ac{RGB-D-T} images from the Bardenas dataset. The learned weights from this pre-training were then used to train the LAENTIEC dataset. During this stage, the first two network layers were frozen, and training was performed for 20 epochs using LAENTIEC images. Table \ref{tab:omniunet-network-metrics} presents the per-class pixel accuracy, as well as overall metrics, including total pixel accuracy (excluding the void class for more realistic results), mean class pixel accuracy, and mean class intersection over union. Additionally, Figures \ref{fig:omniunet-evaluation}a and \ref{fig:omniunet-evaluation}b show the segmentation masks produced by the network with both the Bardenas and LAENTIEC datasets alongside their corresponding color and thermal inputs.

\begin{figure}
\centering
\captionsetup{width = 1\linewidth}
\includesvg[width= 0.48\textwidth]{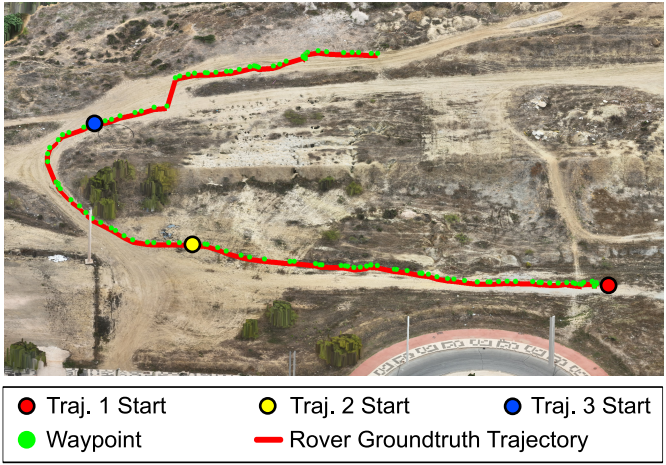}
\caption{\small Trajectories followed by the RAT platform during field tests at the LAENTIEC facility, detailing both autonomous and teleoperated segments.}
\label{fig:rat-laentiec-map}
\end{figure}

\begin{table}[htbp]
\centering
\setlength{\tabcolsep}{3pt}
\renewcommand{\arraystretch}{1.5}
\footnotesize
\captionsetup{width = 1\linewidth}
\caption{\small Dataset statistics and metrics (\%) of the OmniUnet network evaluated on the RUGD, Bardenas, and LAENTIEC datasets. Classes with dashes (-) were not present or not evaluated per class.}
\label{tab:omniunet-network-metrics}
\renewcommand{\arraystretch}{1.2} 
\begin{tabular}{lccc}
\toprule
\textbf{Metric/Class} & \textbf{RUGD} & \textbf{Bardenas} & \textbf{LAENTIEC} \\ 
\midrule
\multicolumn{4}{l}{\textit{Dataset Statistics}} \\
\midrule
Total Images      & 7,435 & 1,140 & 310 \\
Training          & 5,948 & 950   & 248 \\
Validation        & 1,487 & 190   & 62  \\
Number of Classes & 25    & 8     & 6   \\
\midrule
\multicolumn{4}{l}{\textit{Class Pixel Accuracy}} \\
\midrule
Void        & - & 90.48 & 94.16 \\
Compact     & - & 77.44 & 78.68 \\
Grass       & - & 39.04 & 82.72 \\
Bedrock     & - & 28.30 & -     \\
Sandy       & - & 26.57 & -     \\
Gravel      & - & 51.83 & 27.04 \\
Rock        & - & 18.40 & -     \\
Bush/Tree   & - & 40.68 & 41.20 \\
Sky         & - & -     & 94.25 \\
\midrule
\multicolumn{4}{l}{\textit{Average Metrics}} \\
\midrule
Total PA    & \textbf{92.58} & \textbf{80.37} & \textbf{92.85} \\
Mean PA     & 39.72          & 46.59          & 69.68          \\
Mean IoU    & 34.90          & 38.77          & 63.13          \\
\bottomrule
\end{tabular}
\end{table}

\begin{figure}
\centering
\includegraphics[width= 0.49\textwidth]{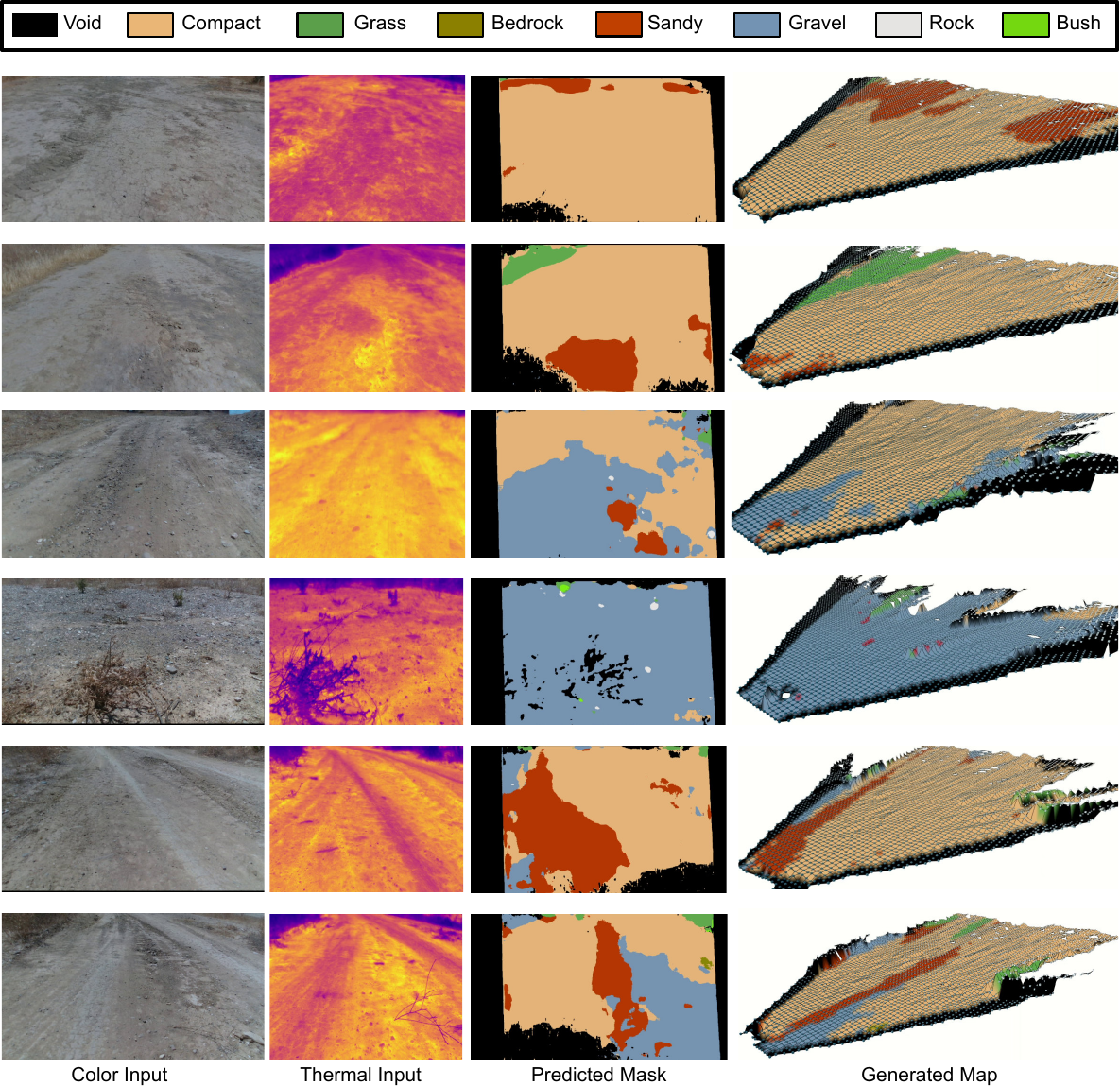}
\caption{\small Visual outputs from the PRISM system during field operations: (a) input RGB image, (b) input thermal image, (c) predicted terrain segmentation mask, and (d) the final 3D traversability map utilized for navigation.}
\label{fig:fieldtest-mapping}
\end{figure}

Regarding the quantitative results, the network achieved a baseline pixel accuracy of \SI{92.58}{\percent} on the RUGD dataset. For the multimodal inputs, both OmniUnet models (trained on Bardenas and LAENTIEC images) achieved a correct pixel classification rate exceeding \SI{80}{\percent}. In the Bardenas dataset, the model demonstrated strong performance in identifying surfaces relevant to safe navigation, successfully detecting compact terrain (the safest to traverse) at \SI{77.44}{\percent} and gravel at \SI{51.83}{\percent}. For obstacle detection, the model identified bushes \SI{40.68}{\percent} of the time; however, rock detection was more difficult, dropping to \SI{18.40}{\percent} likely due to the variability in rock thermal signatures.

\subsection{Experiments}

\begin{table}[t]
\centering
\setlength{\tabcolsep}{3pt} 
\renewcommand{\arraystretch}{1.5} 
\captionsetup{width = 1\linewidth}
\footnotesize	
\caption{Performance metrics and time distribution for the three field test trajectories executed by the RAT rover.}
\begin{tabular}{ccccccc}
\toprule
Traj. & Dist.\ (m) & Gen.\ maps & Stops & Test dur.\ (s) & Motion (s) & Static (s) \\
\midrule
1 & 71 & 91 & 13 & 1881 & 806  & 1075 \\
2 & 67 & 52 & 0  & 1084 & 1084 & 0    \\
3 & 53 & 68 & 9  & 1433 & 633  & 800  \\
\bottomrule
\end{tabular}
\label{tab:rat-fieldtest}
\end{table}

We evaluated the proposed PRISM architecture through a series of field tests with the RAT rover at UMA's LAENTIEC facility. Developed in ROS~2 and accessible on GitHub\footnote{\url{https://github.com/spaceuma/MultimodalNavigation}}, the system constructs traversability maps by integrating DEM data with OmniUnet semantic segmentations. For these experiments, we selected model weights trained on the Bardenas dataset because it closely resembles the summer conditions presented at the test site. Operating at peak CPU capacity, the perception hardware completed the full map generation cycle in approximately \SI{20}{\second} from image capture, while the Terrain Segmenter alone averaged \SI{673}{\milli\second} per prediction. Ultimately, the setup successfully projected traversability maps up to \SI{10}{\meter} ahead of the rover at a resolution of \SI{0.1}{\meter}.

The experimental route comprised three distinct segments (Figure~\ref{fig:rat-laentiec-map}) executed under two operational modes. Trajectories 1 and 3 were navigated autonomously for a combined distance of \SI{124}{\meter}. In this mode, an initial path was computed via the NAV2 interface using a global DEM acquired by a drone. Every \SI{5}{\meter}, the rover paused (indicated by green markers in Figure~\ref{fig:rat-laentiec-map}) to allow the GNC computer to receive an updated local traversability map, integrate it into the global map, and replan the route. Conversely, Trajectory~2 involved a \SI{67}{\meter} teleoperated transition between map zones using a gamepad, during which local maps continued being generated. Table~\ref{tab:rat-fieldtest} summarizes the key metrics from these runs, and Figure~\ref{fig:fieldtest-mapping} presents the resulting PRISM traversability maps. A video recording of the field tests is publicly accessible\footnote{\url{https://youtu.be/6vFK4U6ZmFI}}.

\section{Conclusions and future work}
\label{sec:conclusions}

This study introduced PRISM, a multimodal perception system for terrain segmentation and traversability mapping using RGB-D-T data. At its core, we implemented OmniUnet, a novel transformer-based architecture validated on two new labeled datasets from the Bardenas semi-desert and LAENTIEC facility. OmniUnet demonstrated robust performance, achieving total pixel accuracy exceeding \SI{80}{\percent} on the Bardenas and reaching \SI{92.85}{\percent} on the LAENTIEC dataset. The integration of thermal imagery proved essential for distinguishing surfaces with similar visual appearances but differing thermal inertia, such as compact and sandy soils. We successfully validated the PRISM pipeline through field experiments with the RAT platform, demonstrating efficient, autonomous navigation on resource-constrained hardware. Future work will enhance segmentation robustness by integrating specialized obstacle detection modules and adopting thermal stereo imaging to ensure reliable depth estimation independent of ambient lighting conditions.

\section*{ACKNOWLEDGMENTS}
This work was supported by the Spanish national government project no. PID2024-160373OB-C21, entitled "Hybrid Perception and Navigation for Planetary Exploration".
Additionally, field testing was partially supported by the European Space Agency under activity no.\
4000140043/22/NL/GLC/ces.

\bibliographystyle{IEEEtran}
\bibliography{references}

\vfill
\end{document}